\documentclass[10pt,twocolumn,letterpaper]{article}

\usepackage{cvpr}
\usepackage{times}
\usepackage{epsfig}
\usepackage{graphicx}
\usepackage{amsmath}
\usepackage{amssymb}
\usepackage{multirow}
\usepackage{setspace}
\usepackage{booktabs}
\usepackage[accsupp]{axessibility}
\usepackage{lipsum}
 
\newcommand\blfootnote[1]{%
  \begingroup
  \renewcommand\thefootnote{}\footnote{#1}%
  \addtocounter{footnote}{-1}%
  \endgroup
}

\newcommand{\tabincell}[2]{\begin{tabular}{@{}#1@{}}#2\end{tabular}}

\newcommand{\figref}[1]{Fig.~\ref{#1}}

\newcommand{\tabref}[1]{Tab.~\ref{#1}}

\begin{document}

%%%%%%%%% TITLE
\title{Modeling Spoof Noise by De-spoofing Diffusion and its Application in Face Anti-spoofing}

\author{Bin~Zhang\textsuperscript{1,2}, Xiangyu~Zhu\textsuperscript{3,4,*}, Xiaoyu~Zhang\textsuperscript{1,2}, Zhen~Lei\textsuperscript{3,4,5}\\
\textsuperscript{1}Institute of Information Engineering, Chinese Academy of Sciences\\
\textsuperscript{2}School of Cyber Security, University of Chinese Academy of Sciences\\
\textsuperscript{3}State Key Laboratory of Multimodal Artificial Intelligence Systems,\\ Institute of Automation, Chinese Academy of Sciences\\
\textsuperscript{4}School of Artificial Intelligence, University of Chinese Academy of Sciences\\
\textsuperscript{5}Centre for Artificial Intelligence and Robotics, Hong Kong Institute of Science \& Innovation,\\Chinese Academy of Sciences\\ 
{\tt\small \{zhangbin1998, zhangxiaoyu\}@iie.ac.cn, \{xiangyu.zhu, zlei\}@nlpr.ia.ac.cn}
%\textsuperscript{1}\texttt{author1@example.com},
%\textsuperscript{2}\texttt{author2@example.com}
}
\maketitle
\thispagestyle{empty}
\blfootnote{* corresponding author.}

%%%%%%%%% ABSTRACT
\begin{abstract}
Face anti-spoofing is crucial for ensuring the security and reliability of face recognition systems. Several existing face anti-spoofing methods utilize GAN-like networks to detect presentation attacks by estimating the noise pattern of a spoof image and recovering the corresponding genuine image. But GAN's limited face appearance space results in the denoised faces cannot cover the full data distribution of genuine faces, thereby undermining the generalization performance of such methods. In this work, we present a pioneering attempt to employ diffusion models to denoise a spoof image and restore the genuine image. The difference between these two images is considered as the spoof noise, which can serve as a discriminative cue for face anti-spoofing. We evaluate our proposed method on several intra-testing and inter-testing protocols, where the experimental results showcase the effectiveness of our method in achieving competitive performance in terms of both accuracy and generalization.
\end{abstract}

%%%%%%%%% BODY TEXT
\section{Introduction}
With the development of embedded devices and smart devices, face authentication systems have become widely employed in our daily life. From unlocking portable devices to mobile payment, people are willing to use convenient and secure ways to access their identity information. However, hackers exploit the vulnerability of face recognition models to design various presentation attacks in order to deceive identity authentication systems. Print and replay attacks are the most common identity attacks, as no-cost access to the human face by simply presenting the target face from a printed photo or a display device. Therefore, the eagerness for securing face authentication systems from challenging presentation attacks promotes the techniques of face anti-spoofing. 

Although traditional CNNs have demonstrated impressive results in the realm of computer vision, they still confront challenges in face anti-spoofing, particularly regarding generalization performance~\cite{cai2022learning,shao2019regularized}. This is because the cues for liveness detection often reside in fine-grained information that is not related to semantic information. In recent years, numerous methods have emerged that utilize estimating auxiliary signals from RGB data to uncover internal cues between spoof and genuine images, such as depth maps~\cite{atoum2017face,liu2018learning}, rPPG~\cite{liu2018learning}, color texture~\cite{boulkenafet2015face}, distortion~\cite{wen2015face} and reflectance maps~\cite{pinto2020leveraging}, and have shown the potential to achieve better generalization performance. By estimating auxiliary information, it is possible to highlight the invisible spoofing cues and improve the generalization performance of the face anti-spoofing model.

\begin{figure*}[htp]
\begin{center}
%\fbox{\rule{0pt}{2in} \rule{0.9\linewidth}{0pt}}
   \includegraphics[width=0.9\linewidth]{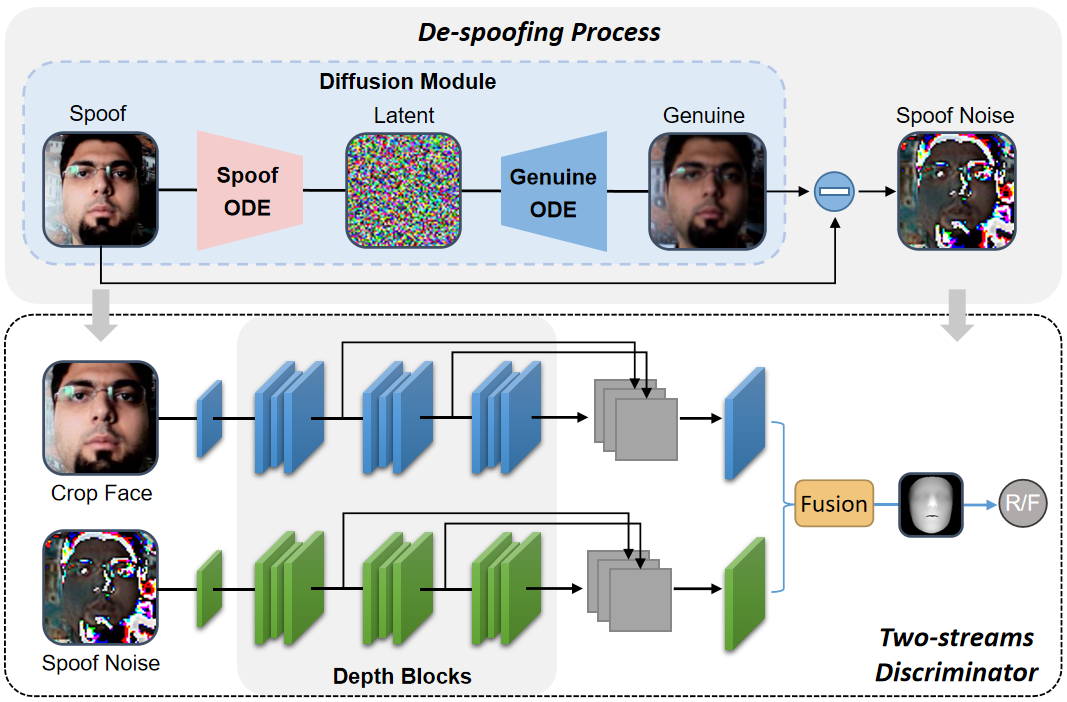}
\end{center}
   \caption{Our proposed de-spoofing diffusion method decomposes a face image into its genuine counterpart and spoof noise pattern. The noise pattern, along with the RGB input, is then utilized to train the discriminator for the accurate detection of presentation attacks.}
\label{fig:network}
\end{figure*}

The secondary imaging process that recaptures a spoof image inevitably brings along subtle noise. This noise is caused by the artifact signals that are captured by the camera sensor, including factors like abnormal light reflections, non-rigid deformation, medium types, and camera parameters. In this paper, we aim to disentangle a spoof image into the genuine image and spoof noise pattern which serves as auxiliary cues for face anti-spoofing. ~\figref{fig:network} shows the pipeline of our proposed method with two steps: face de-spoofing and spoof detection. Our idea of De-Spoofing is similar to classical De-X problems, which attempt to remove the degradation effect or artifacts from the image. But in De-X problems, the degraded samples are obtained by adding noise to the original samples, such as de-noising~\cite{buades2005non,lefkimmiatis2017non}, de-mosaicing~\cite{li2005demosaicing}, and super-resolution~\cite{tai2017image}. In contrast, since the spoof images are captured from spoof mediums rather than generated from genuine images, there is no corresponding ground truth with the same identity information for these spoof images. Jourabloo et al.~\cite{jourabloo2018face} first propose to use GAN to estimate the real images from the corresponding spoof images under the supervision of all the genuine images in the training set. Then, the noise image is obtained by subtracting the estimated real image from the spoof image. Subsequently, many GAN-based noise modeling methods~\cite{liu2020disentangling,xu2021identity,wang2022domain} have been proposed for face anti-spoofing. The noise patterns in face anti-spoofing exhibit multiple distributions, but the adversarial loss does not have the incentive to cover the entire data distribution, which is a common downside of GANs~\cite{metz2016unrolled}, causing a reduction in the generalization performance of these GAN-based methods. Meanwhile, these methods are also difficult to optimize~\cite{brock2018large}, collapsing without carefully selected hyperparameters and regularizers.  

Recently, diffusion models showcase impressive generative capabilities which have been applied to a wide variety of generative modeling tasks, such as image generation~\cite{ho2020denoising,dhariwal2021diffusion,song2020denoising}, image editing~\cite{meng2021sdedit,avrahami2022blended}, style transfer~\cite{su2022dual,liu2022flow}, super-resolution~\cite{rombach2022high,saharia2022image}, etc. DDPMs~\cite{ho2020denoising} are among the most widely used diffusion models, which learn to reverse the degradation process of the training data structure by gradually adding Gaussian noise at different scales. Our method utilizes a diffusion model to learn the distribution of genuine faces and disentangles the spoof noise patterns from spoof faces by reversing the degradation process. In conditional DDPMs~\cite{dhariwal2021diffusion}, an additional presentation attacks detector is required to steer the reverse process of diffusion when transferring spoof images to the distribution of genuine images. The detection performance of the model ultimately depends on the effectiveness of the initial guide classifier, which does not address the generalization issue. Inspired by the work of style transfer~\cite{su2022dual}, we propose to utilize the DDIM~\cite{song2020denoising} model to separate the degradation process of the spoof images and the reverse process of the genuine images into two independent stages. Each stage learns the distribution of either the face anti-spoofing dataset or the distribution of genuine faces, respectively. Notably, our approach does not necessitate the guidance of a classifier and incentivizes learning the full data distribution, which significantly enhances its generalization ability.

Overall, our core contributions are summarized as follows:

1) It introduces a novel approach for modeling spoof noise by designing a de-spoofing diffusion model, which is employed as auxiliary cues for the detection of presentation attacks.

2) We propose a de-spoofing network that successfully transfers spoof images into the corresponding genuine images without a guide classifier, and the noise pattern can then be extracted.

3) Our method not only outperforms the state-of-the-art methods on the intra-testing of CASIA-MFSD, Replayattack, and OULU-NPU datasets, but also exhibits superior results on cross-dataset testing with CASIA-MFSD and Replay-attack.

%------------------------------------------------------------------------
\section{Related Works}
\subsection{Deep Learning Methods}
Deep learning techniques have made great strides in the field of computer vision\cite{voulodimos2018deep}. Benefited from the release of large-scale face anti-spoofing dataset\cite{chingovska2012effectiveness,wen2015face,boulkenafet2017oulu,zhang2012face}, the advanced CNN architectures~\cite{he2016deep,chollet2017xception} and regularization techniques have been widely used in face anti-spoofing. In the field of face anti-spoofing, CNN was initially used as a pre-trained feature extractor, followed by SVM for the classification of the extracted features~\cite{yang2014learn}. Subsequent methods utilized cross-entropy loss to transform the face anti-spoofing problem into an end-to-end binary classification CNN. For example, Rehman et al.~\cite{rehman2017deep} propose an end-to-end deep face anti-spoofing method using CNN for feature representation with a linear. To increase the model's prediction confidence on hard samples, Chen et al.~\cite{chen2021camera} adopt the binary focal loss to enlarge the margin between genuine and spoof samples, thereby achieving better discrimination for difficult samples. Although these models have achieved excellent results in intra-testing, their effectiveness is limited in cross-dataset testing due to the limited amount and diversity of
the training data. To address the negative effects of overfitting, some methods fine-tune the ImageNet-pretrained models for face anti-spoofing. ResNet~\cite{pinto2020leveraging}, VGG~\cite{lucena2017transfer}, and Transformer~\cite{george2021effectiveness} are commonly used as backbones for generalized methods against invisible attack types. Subsequent studies have shown that binary supervision is prone to learning disloyal patterns, such as identity information, collection environment, and fraud media boundaries, which weakens the model's generalization ability. Meanwhile, pixel-level supervision has been shown to provide fine-grained information that can enhance detection performance. Depth information~\cite{atoum2017face,liu2018learning,yu2020searching} is the most commonly used supervision label in such methods, which enforces the detection models to predict the actual depth maps for genuine samples while zero depth maps for the spoof ones. Compared to binary supervision, depth is more informative than binary labels since it indicates one of the fundamental differences between genuine and spoofing faces. Atoum et al.~\cite{atoum2017face} propose a patch and depth-based CNNs, in which the patch branch is utilized to force the network focus on local information, and the depth branch is incorporated to capture global information. Instead of directly predicting the depth map, Liu et al.~\cite{liu2018learning} leverage pseudo-depth labels to guide the training of the network. To further improve the fine-grained intrinsic feature representation capacity, Yu et al.~\cite{yu2020searching} replace vanilla convolution in DepthNet~\cite{liu2018learning} with central difference convolution to form the CDCN architecture.

%-------------------------------------------------------------------------
\subsection{Hybrid Methods Combining Auxiliary Features and Deep Features}
Previous methods for face anti-spoofing based on CNNs are mainly trained and evaluated using RGB input alone. However, these approaches have encountered several challenges, including overfitting and limited generalization ability. To address these issues, recent studies have explored the incorporation of auxiliary information into CNN-based models, which can effectively enhance the performance of face anti-spoofing models. Early handcrafted features~\cite{cai2019learning,chen2019attention,feng2016integration} are utilized as auxiliary information in some methods for the input of detection models, which are proven to be discriminative in distinguishing genuine samples from presentation attacks. For example, Cai et al.~\cite{cai2019learning} adopt multi-scale color LBP features as local texture descriptors, followed by a cascaded random forest for semantic representation. Then, Chen et al.~\cite{chen2019attention} propose an attention-based fusion framework that utilizes a two-stream CNN to fuse two complementary color spaces of RGB and MSR. In addition to static features, dynamic features across temporal frames are also effective CNN inputs. Feng et al.~\cite{feng2016integration} propose to train a multi-layer perceptron using dense optical flow-based motions extracted from video frames, which can reveal anomalies in print attacks. Liu et al.~\cite{liu2018learning} also propose a method by combining spatial and temporal features from depth maps and spatio-temporal rPPG signals. Moreover, researchers attempt to capture dynamic features by analyzing the facial movements in a sequence of frames, such as the motion of eye blinks~\cite{pan2007eyeblink}, mouth and lip~\cite{kollreider2007real}. Although these methods can effectively discern print attacks, they become vulnerable to replay attacks or print attacks with eye/mouth cuts.
Furthermore, physical signals from spoof mediums are another effective and interpretable source of auxiliary information. For example, Bian et al.~\cite{bian2022learning} present a framework that learns multiple generalizable cues from moiré patterns, reflection artifacts, facial depth, and the boundary of the spoofing medium. Reflectance of the face image is another commonly used cues for face anti-spoofing, as it reflects the material differences between genuine and spoof faces, particularly for print and replay attacks. Pinto et al.~\cite{pinto2020leveraging} utilize reflectance, depth, and albedo maps to detect presentation attacks. Integrating auxiliary features with deep neural networks is a promising direction in face anti-spoofing, as it can provide good explanatory and generalization performance.

%-------------------------------------------------------------------------

\subsection{Spoof Noise-based Methods}
Since the spoof images are obtained from secondary imaging by presenting the target faces from spoof mediums, artifacts noise will be inevitably introduced to the genuine images. The process of de-spoofing is similar to the classic De-X problem, such as de-noising, de-blurring, and de-mosaicing, both of which have the common goal of removing the degradation effect or artifacts from the image. For instance, typical image de-noising methods assume the presence of additive Gaussian noise, and various techniques have been proposed to leverage the intrinsic similarity within the image, such as non-local filters~\cite{buades2005non} or CNNs~\cite{lefkimmiatis2017non}. While face de-spoofing problem aims to estimate the noise pattern of a spoof image to detect presentation attacks. Liu et al.~\cite{jourabloo2018face} first propose to estimate the genuine image from the corresponding spoof image under the supervision of all the genuine images. Xu et al.~\cite{xu2021identity} further introduced the identity information of the target person and proposed a metric learning module to constrain the genuine images generated from the spoof images to enhance the de-spoofing effect. Similarly, Liu et al.~\cite{liu2020disentangling} also design a generic anti-spoofing model to estimate the spoof traces and enable the reconstruction of genuine faces. To further enhance the domain generalization performance, Wang et al.~\cite{wang2022domain} propose to separate the content features and spoof noise patterns by utilizing style transfer. Cai et al.~\cite{cai2022learning} propose a bi-level optimization method to extract noise patterns in an end-to-end data-driven way. With the advancement of spoof materials, the visual differences have been reduced to a low magnitude. Therefore, analyzing the spoof noise in imaging can further help us better understand what the patterns of the classifier’s decision are based upon.

%-------------------------------------------------------------------------
\begin{figure*}[htp]
\begin{center}
%\fbox{\rule{0pt}{2in} \rule{0.9\linewidth}{0pt}}
   \includegraphics[width=0.95\linewidth]{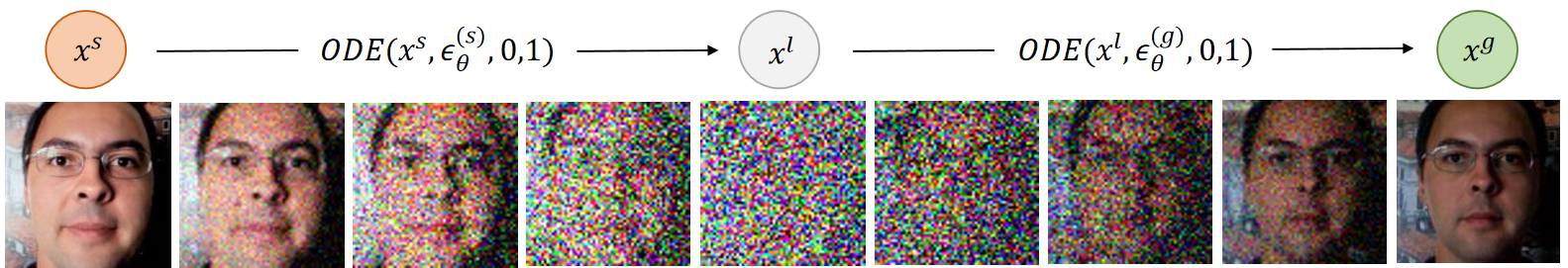}
\end{center}
   \caption{Our method leverage two ODEs for image de-spoofing. Given a source image $x^{s}$ (e.g. a spoof image or genuine image), the first ODE runs in the forward direction to convert it to the latent $x^{l}$, while the target, reverse ODE then constructs the target genuine image $x^{g}$.}
\label{fig:DDIB}
\end{figure*}

\subsection{Diffusion Model}
Diffusion models have emerged as the new state-of-art family of generative models that learn to reverse a process that gradually degrades the training data by adding noise at different scales. Diffusion models break the long-standing dominance of generative adversarial networks~\cite{GAN2014} in image synthesis tasks, such as image generation, image editing, style transfer, super-resolution, etc. Among them, Denoising diffusion probabilistic models (DDPMs) are the most widely used diffusion models, which makes use of two Markov chains: a forward chain that perturbs data to noise, and a reverse chain that converts noise back to data. To precisely control the generation target of the diffusion model, the conditional diffusion model proposed by Dhariwal et al.~\cite{dhariwal2021diffusion} utilizes a classifier to correct the reverse results of the diffusion model. Despite these powerful performances, the diffusion models are still not efficient enough as they require multiple network evaluations during inference. Subsequently, Song et al.~\cite{song2020denoising} propose DDIM, which can reversely generate noised images and accelerate the overall generation process. Inspired by DDIM, Su et al.~\cite{su2022dual} utilize the particular parameterization of the diffusion process to transfer images from one domain to another without concurrent access to both datasets. While certain studies~\cite{blasingame2023diffusion,damer2023mordiff} have employed diffusion models to generate stronger attacks, there is still a lack of exploration in the domain of face anti-spoofing.

\section{Approach}
In this section, we propose a novel face anti-spoofing method based on the state-of-the-art generative technique of diffusion models, which is shown in ~\figref{fig:network}. A de-spoofing diffusion model is introduced to acquire spoof noise patterns for presentation attacks. Furthermore, a two-stream fusion network is proposed to capture the spoof cues from RGB input and noise patterns.

\subsection{Cues in Spoof Noise}
During the secondary imaging of spoof images, noise patterns are implicitly encoded into the genuine image by the camera sensor, including color distortion, moiré pattern, medium edge, medium types, abnormal light reflections, non-rigid deformation, illumination conditions, etc. Under the influence of these variable combinations, the spoof images will undergo degradation compared to the genuine images. Based on this observation, our de-spoofing method aims to disentangle the spoof sample into their genuine composition and spoof noise pattern, providing fine-grained evidence to support the model’s decision.

Let the domain of spoof images be denoted as $S \in \mathbb{R}^{n\times n\times 3}$ and genuine images as $G \in \mathbb{R}^{n\times n\times 3}$, where $n$ is the image size. Our aim is to transfer an input face to the genuine domain and obtain the noise pattern by:
\begin{equation}\label{equ-noise}
    N = \left \| \mathbf{I} - \mathbf{\hat{I}}   \right \| s.t. \mathbf{I}\in \left ( S\cup G  \right )  \ and \  \mathbf{\hat{I}} \in G.
\end{equation}

where $\mathbf{I}$ is the input image, $\mathbf{\hat{I}}$ is the target face in the genuine domain, and $N$ is defined as the spoof noise. But in de-X problems, the degraded images are generated from the input images which are treated as the intuitive ground truth. While in the de-spoofing problem, the spoof images are not directly generated from genuine images but captured from a spoof medium, resulting in spoof images without corresponding genuine images. To solve this issue, we propose to utilize the powerful learning and generation capabilities of the diffusion model to transfer a spoof sample into the corresponding genuine domain sample.

\subsection{De-spoofing Diffusion}
Typically, diffusion models contain two steps: the forward step that smoothly perturbs input data $x_{0}$ to noise $x_{t}$ which is a typically hand-designed process, and the reverse step that reverses a gradual noising process back to data $x_{0}$. In other words, diffusion models start with noise $x_{t}$ and produces gradually less-noisy samples $x_{t-2}$, $x_{t-2}$ ..., until reaching a final sample $x_{0}$. Each timestep $t$ corresponds to a certain noise level, and $x_{t}$ can be thought of as a mixture of a signal $x_{0}$ with some noise (e.g., Gaussian noise). To parameterize the DDPMs, UNet~\cite{ronneberger2015u} $\epsilon _{\theta } (x_{t},t)$ is utilized to predict the noise component of a noisy input sample $x_{t}$. During the training of these models, each sample in a minibatch is produced by randomly drawing a data sample $x_{0}$, a timestep $t$, and noise $\theta$, which together give rise to a noised sample  $x_{t}$. The objective of the training is then optimized by minimizing ${\left \|  \epsilon _{\theta } (x_{t},t)   -   \epsilon   \right \| } ^2$. 

However, the sampling process of DDPM relies on independent distribution rather than joint distribution~\cite{song2020denoising}, which leads to low efficiency of DDPM generation. Subsequently, DDIM formulates an alternative non-Markovian noising process that has the same forward marginals as DDPM but allows producing different reverse samplers by changing the variance of the reverse noise. This demonstrates that the diffusion process can be represented by a deterministic probability flow ordinary differential equation, which carries the same marginal density as the diffusion process along the entire trajectory. We define the symbol $ODE(x_{t_{0}};\epsilon _{\theta },t_{0},t_{1})$ to denote the mapping from $x_{t_{0}}$ to $x_{t_{1}}$, where $t_{0},t_{1} \in [0,1]$. By leveraging the connection between the score-based generative model (SGM) and Schrödinger Bridge Problem (SBP)~\cite{su2022dual}, we can train separate diffusion models for the source and target domains and perform image-to-image transformations. The de-spoofing process based on the diffusion model can be illustrated in ~\figref{fig:DDIB}. To obtain the latent code from the union of the genuine domain and spoof domain, we generate the noise image by the ordinary differential equation:
\begin{equation}\label{equ-ODE1}
    x^{l} = ODE(x^{s};{\epsilon _{\theta }}^{(s)},0,1),
\end{equation}
where $x^{l}$ is the latent code, ${\epsilon _{\theta }}^{(s)}$ is the diffusion model trained on a complete face anti-spoofing dataset, and $x^{s} \in (S\cup G)$. Further, the latent code is transformed into the genuine domain by:
\begin{equation}\label{equ-ODE2}
    x^{g} = ODE(x^{l};{\epsilon _{\theta }}^{(g)},1,0),
\end{equation}
where $x^{g}$ is the genuine image corresponding to the input image $x^{s}$, ${\epsilon _{\theta }}^{(g)}$ is the diffusion model only trained on a genuine face data. Since the spoof images are regarded as the addition of noise and the genuine counterparts, the noise patterns can be expressed as:
\begin{equation}\label{equ-Noise}
    N = \left \| x^{s} - x^{g}   \right \|.
\end{equation}

\subsection{Noise Fusion Model}
To effectively capture the spoofing cues from both the RGB input and extracted noise patterns, we propose a two-stream fusion network as depicted at the bottom of ~\figref{fig:network}. Since face anti-spoofing models benefit from pixel-level supervision, we adopt a three-layer structure in the form of DepthNet~\cite{liu2018learning} as the backbone of our two-stream fusion network. The first branch of the model receives RGB input and applies center crop preprocessing to adjust the proportion of background information in the image. The second branch inputs the noise information from the complete image to assist the model's decision. Finally, the results of the last layer of each branch are fused to obtain the depth prediction map. To enhance the model's ability to extract fine-grained invariant information, we employ Central Difference Convolution~\cite{yu2020searching} in conjunction with vanilla convolution. For the depth-supervised network, mean square error loss $L_{MSE}$ is utilized for pixel-wise supervision, and contrastive depth loss $L_{CDL}$~\cite{wang2018exploiting} is considered to improve the generalization of our model.
\begin{equation}\label{equ-CDCN}
   L_{overall} = L_{MSE} + L_{CDL}.
\end{equation}

\section{Experiments}
\subsection{Datasets $\&$ Metrics}
\noindent \textbf{Databases.} Four datasets CASIA-MFSD~\cite{zhang2012face}, Replay-Attack~\cite{chingovska2012effectiveness}, OULU-NPU~\cite{boulkenafet2017oulu}, and SiW~\cite{liu2018learning} are used to evaluate our proposed network with print and replay attacks. 

%OULU-NPU and SiW are high-resolution datasets, while CASIA-MFSD and Replay-attack are low-resolution datasets.

%OULU-NPU and SiW are high-resolution databases, containing four and three protocols to validate model generalization, respectively. CASIA-MFSD and Replay-Attack are databases that contain low-resolution videos, which are used for intra-testing and inter-testing.

\noindent \textbf{Performance Metrics.} The performance of our proposed method is assessed through three basic metrics on the OULU-NPU and SiW dataset, including the Attack Presentation Classification Error Rate (APCER), the Bona fide Presentation Classification Error Rate (BPCER), and the Average Classification Error Rate (ACER) for a fair comparison. Then the Equal Error Rate (EER) and Half Total Error Rate (HTER) are adopted in the testing of CASIA-MFSD and Replay-attack, respectively.

\subsection{Implementation Details}\label{Implementation}
Our proposed method is implemented with the Pytorch 1.12.0 framework.

\noindent \textbf{The training of diffusion model}. We first train a denoising diffusion probabilistic model on the face anti-spoofing dataset. Then we train another denoising diffusion model only on genuine face data. We set $T$ = 1000 for all training experiments during the sampling process. While DDIM with 50 steps is adopted for the de-spoofing process.

\noindent \textbf{The training of two-stream fusion network}. Our proposed two-stream fusion network employs a Halfway fusion strategy to combine the results of the two branches. Additionally, on the OULU-NPU and SiW datasets, we utilize score fusion to further enhance the intra-testing performance of the model. The depth supervision map of the network is generated by 3DDFA~\cite{zhu2017face}. Then, the two-stream fusion network is trained with the Adam optimizer and the initial learning rate and weight decay are 1e-4 and 5e-5. We train the model with a maximum of 1000 epochs and adopt a step learning rate strategy, which decays every 500 training steps by a factor of 0.1. The batch size is 64 on four TITAN XP GPUs.

\subsection{Ablation Study}
\noindent \textbf{Impact of the Noise Patterns.} In this section, ablation studies are conducted on cross-dataset testing of CASIA-MFSD and Replay-attack. Specifically, we design four experiments to investigate the effectiveness and generalization of our method under different conditions. First, we employ a single-stream discrimination network that takes only RGB images or noise patterns as input. While the third experiment consists of training a two-stream fusion network with only RGB data available. In the last experiment, we replaced the second branch of the two-stream network with noisy inputs for training. The experimental results shown in ~\tabref{tab:RC-ab} demonstrate that our proposed two-stream fusion network slightly improves the performance of the single-stream network. Moreover, by integrating noise patterns into the fusion network, we can further explore spoof cues that help the model's decision.

\noindent \textbf{Impact of the Timesteps.} We conduct the experiment that sampling for different timesteps with DDIM, which affects the quality of denoising samples. In the case of timesteps 25, 50, and 100, the EER($\%$) of the intra-testing of CASIA-MFSD are 0.83, 0.37, and 0.56.

\begin{table}
\begin{spacing}{1.2}
    \begin{center}
        \caption{Verify the effectiveness and generalization of our de-spoofing method based on the cross-data testing protocol. The evaluation metric is HTER($\%$). A lower HTER is better.}
        \label{tab:RC-ab}
        \resizebox{\linewidth}{!}{
        \begin{tabular}{|c|c|c|c|c|} 
            \hline 
            \multicolumn{1}{|c|}{\multirow{2}{*}{Inputs}} & Train & Test & Train & Test  \\ \cline{2-5}
            & \tabincell{l}{CASIA- \\ MFSD} & \tabincell{l}{Replay \\ Attack} & \tabincell{l}{Replay \\ Attack} &  \tabincell{l}{CASIA- \\ MFSD} \\
            \hline
            RGB & \multicolumn{2}{c}{13.0} & \multicolumn{2}{|c|}{36.1} \\
            \hline
            Noise & \multicolumn{2}{c}{18.1} & \multicolumn{2}{|c|}{39.4} \\ %37.4
            \hline
            RGB and RGB & \multicolumn{2}{c}{12.1} & \multicolumn{2}{|c|}{35.7} \\
            \hline
            \textbf{RGB and Noise} & \multicolumn{2}{c}{\textbf{9.8}} & \multicolumn{2}{|c|}{\textbf{34.4}} \\
            \hline
        \end{tabular}}
    \end{center}
\end{spacing} 
\end{table}

\begin{table}
  \begin{center}
  {\tiny
  \caption{The results of intra-testing on CASIA-MFSD and Replay-attack. The performance metric are EER($\%$) for CASIA-MFSD, and HTER($\%$) for Replay-attack. “$\downarrow$” means the lower the better.}
  \label{tab:intra-CR}
  \setlength{\arrayrulewidth}{0.1pt}
  \renewcommand\arraystretch{1.2}
  \resizebox{\linewidth}{!}{
  \begin{tabular}{|c|c|c|}
    \hline
    \multicolumn{1}{|c|}{Method} & \multicolumn{1}{c|}{CASIA-MFSD} & \multicolumn{1}{c|}{Replay-attack} \\
    \cline{2-3}
    & \multicolumn{1}{c|}{EER($\%$)$\downarrow$} & \multicolumn{1}{c|}{HTER($\%$)$\downarrow$} \\
    \hline
    MIQF~\cite{chang2022face} & 12.7 & 5.38 \\
    \hline
    CNN~\cite{yang2014learn} & 7.4 & 2.1 \\
    \hline
    Color-LBP~\cite{boulkenafet2015face} & 6.2 & 2.9 \\
    \hline
    SfSNet~\cite{pinto2020leveraging} & 3.3 & 3.1 \\
    \hline
    Attention~\cite{chen2019attention} & 3.15 & 0.39 \\
    \hline
    CDCN~\cite{yu2020searching} & 2.9 & 0.38 \\
    \hline
    DANet~\cite{sun2022danet} & 2.5 & - \\
    \hline
    CAMERA~\cite{chen2021camera} & 0.89 & - \\
    \hline
    \textbf{Ours} & \textbf{0.37} & \textbf{0.25} \\
    \hline
  \end{tabular}}}
  \end{center}
\end{table}

\subsection{Intra-Dataset Evaluation}
The intra-testings are carried out on CASIA-MFSD, Replay-attack, OULU-NPU, and SiW datasets. We strictly follow the intra-testing protocols of these datasets to make a fair evaluation.

\begin{table}
    \begin{center}
        \caption{The results of intra-testing on four protocols of OULU-NPU. The performance metric are APCER($\%$), BPCER($\%$) and ACER($\%$). “$\downarrow$” means the lower the better.}
        \label{tab:OULU-NPU}
            \resizebox{\linewidth}{!}{
            \begin{tabular}{|c|c|c|c|c|c|} 
                \hline 
                Prot & Method & APCER($\%$)$\downarrow$ & BPCER($\%$)$\downarrow$ & ACER($\%$)$\downarrow$  \\ 
                \hline 
                \multirow{9}{*}{1} & CAMERA~\cite{chen2021camera} & 3.8 & 2.9 & 3.4 \\ 
                \multirow{9}{*}{ } & STASN~\cite{yang2019face} & 1.2 & 2.5 & 1.9 \\ 
                \multirow{9}{*}{ } & Auxiliary~\cite{liu2018learning} & 1.6 & 1.6 & 1.6 \\ 
                \multirow{9}{*}{ } & FaceDs~\cite{jourabloo2018face} & 1.2 & 1.7 & 1.5 \\
                \multirow{9}{*}{ } & Identity-DS~\cite{xu2021identity} & 1.2 & 1.6 & 1.4 \\
                \multirow{9}{*}{ } & FAS-TD~\cite{wang2018exploiting} & 2.5 & \textbf{0} & 1.3 \\
                \multirow{9}{*}{ } & SpoofTrace~\cite{liu2020disentangling} & 0.8 & 1.3 & 1.1 \\
                \multirow{9}{*}{ } & CDCN~\cite{yu2020searching} & 0.4 & 1.7 & 1.0 \\
                \multirow{9}{*}{ } & DANet~\cite{sun2022danet} & \textbf{0} & 1.6 & 0.8 \\
                \multirow{9}{*}{ } & \textbf{Ours} & 0.4 & 0.8 & \textbf{0.6} \\ 
                \hline
                \multirow{9}{*}{2} & FaceDs~\cite{jourabloo2018face} & 4.2 & 4.4 & 4.3 \\ 
                \multirow{9}{*}{ } & Auxiliary~\cite{liu2018learning} & 2.7 & 2.7 & 2.7 \\
                \multirow{9}{*}{ } & CAMERA~\cite{chen2021camera} & 3.6 & 1.2 & 4.4 \\ 
                \multirow{9}{*}{ } & STASN~\cite{yang2019face} & 4.2 & \textbf{0.3} & 2.2 \\
                \multirow{9}{*}{ } & Identity-DS~\cite{xu2021identity} & 2.5 & 2.9 & 2.2 \\
                \multirow{9}{*}{ } & FAS-TD~\cite{wang2018exploiting} & 1.7 & 2.0 & 1.9 \\ 
                \multirow{9}{*}{ } & SpoofTrace~\cite{liu2020disentangling} & 2.3 & 1.6 & 1.9 \\
                \multirow{9}{*}{ } & DANet~\cite{sun2022danet} & \textbf{0.8} & 2.4 & 1.6 \\
                \multirow{9}{*}{ } & CDCN~\cite{yu2020searching} & 1.5 & 1.4 & 1.5 \\
                \multirow{9}{*}{ } & \textbf{Ours} & 1.0 & 1.1 & \textbf{1.0} \\  
                \hline
                \multirow{9}{*}{3} & FAS-TD~\cite{wang2018exploiting} & 5.9±1.9 & 5.9±3.0 & 5.9±1.0 \\ 
                \multirow{9}{*}{ } & FaceDs~\cite{jourabloo2018face} & 4.0±1.8 & 3.8±1.2 & 3.6±1.6 \\ 
                \multirow{9}{*}{ } & SpoofTrace~\cite{liu2020disentangling} & 1.6±1.6 & 4.0±5.4 & 2.8±3.3 \\
                \multirow{9}{*}{ } & Auxiliary~\cite{liu2018learning} & 2.7±1.3 & 3.1±1.7 & 2.9±1.5 \\
                \multirow{9}{*}{ } & Identity-DS~\cite{xu2021identity} & 3.1±3.0 & 2.4±1.5 & 2.8±2.1 \\ 
                \multirow{9}{*}{ } & STASN~\cite{yang2019face} & 4.7±3.9 & \textbf{0.9±1.2} & 2.8±1.6 \\ 
                \multirow{9}{*}{ } & CAMERA~\cite{chen2021camera} & 3.8±1.3 & 1.1±1.1 & 2.5±0.8 \\
                \multirow{9}{*}{ } & CDCN~\cite{yu2020searching} & 2.4±1.3 & 2.2±2.0 & 2.3±1.4 \\
                \multirow{9}{*}{ } & DANet~\cite{sun2022danet} & 1.8±1.6 & 2.7±1.0 & 2.2±1.2 \\
                \multirow{9}{*}{ } & \textbf{Ours} & \textbf{0.6±0.6} & 1.7±3.0 & \textbf{1.1±1.4} \\  
                \hline
                \multirow{9}{*}{4} & Auxiliary~\cite{liu2018learning} & 9.3±5.6 & 10.4±6.0 & 9.5±6.0 \\ 
                \multirow{9}{*}{ } & FAS-TD~\cite{wang2018exploiting} & 14.2±8.7 & \textbf{4.2±3.8} & 9.2±3.4 \\ 
                \multirow{9}{*}{ } & STASN~\cite{yang2019face} & 6.7±10.6 & 8.3±8.4 & 7.5±4.7 \\
                \multirow{9}{*}{ } & CDCN~\cite{yu2020searching} & 4.6±4.6 & 9.2±8.0 & 6.9±2.9 \\
                \multirow{9}{*}{ } & CAMERA~\cite{chen2021camera} & 5.9±3.3 & 6.3±4.7 & 6.1±4.1 \\
                \multirow{9}{*}{ } & DANet~\cite{sun2022danet} & \textbf{0.8±1.4} & 11.2±8.1 & 6.0±4.9 \\
                \multirow{9}{*}{ } & FaceDs~\cite{jourabloo2018face} & 1.2±6.3 & 6.1±5.1 & 5.6±5.7 \\
                \multirow{9}{*}{ } & Identity-DS~\cite{xu2021identity} & 4.1±3.4 & 6.6±5.5 & 5.4±3.0 \\
                \multirow{9}{*}{ } & SpoofTrace~\cite{liu2020disentangling} & 2.3±3.6 & 5.2±5.4 & \textbf{3.8±4.2} \\
                \multirow{9}{*}{ } & \textbf{Ours} & 3.3±3.4 & 6.6±4.7 & 5.0±3.9 \\  
                \hline
            \end{tabular}
            }
    \end{center}
\end{table}

\begin{table}
    \begin{center}
        \caption{The results of intra-testing on the third protocol of SiW. The performance metric are APCER($\%$), BPCER($\%$) and ACER($\%$). “$\downarrow$” means the lower the better.}
        \label{tab:SiW}
            \resizebox{\linewidth}{!}{
            \begin{tabular}{|c|c|c|c|c|c|} 
                \hline 
                Prot & Method & APCER($\%$)$\downarrow$ & BPCER($\%$)$\downarrow$ & ACER($\%$)$\downarrow$  \\ 
                \hline 
                \multirow{5}{*}{3} & STASN~\cite{yang2019face} & - & - & 12.10±1.50 \\ 
                \multirow{5}{*}{ } & Auxiliary~\cite{liu2018learning} & 8.31±3.81 & 8.31±3.80 & 8.31±3.81 \\ 
                \multirow{5}{*}{ } & 3DPC-Net~\cite{li20203dpc} & 7.50±38.81 & 7.85±38.13 & 7.68±38.50 \\
                \multirow{5}{*}{ } & SDCC~\cite{zhang2021structure} & 3.80±4.30 & \textbf{3.0±2.60} & 3.40±0.90 \\
                \multirow{5}{*}{ } & \textbf{Ours} & \textbf{3.29±1.98} & 3.51±2.0 & \textbf{3.39±1.99} \\ 
                \hline
            \end{tabular}
            }
    \end{center}
\end{table}

\emph{1)Result on CASIA-MFSD and Replay-attack:} In this experiment, we evaluated our proposed method on two datasets, CASIA-MFSD and Replay-Attack, utilizing two different metrics: EER and HTER. The results on CASIA-MFSD and Replay-Attack are shown in ~\tabref{tab:intra-CR}. Specifically, our method achieved an EER of 0.37$\%$ on CASIA-MFSD and an HTER of 0.25$\%$ on Replay-Attack, outperforming the existing methods.

\emph{2)Result on OULU-NPU:}
The OULU-NPU dataset is a widely used benchmark for evaluating the generalization performance of face anti-spoofing methods. In \tabref{tab:OULU-NPU}, we present a comparison of our approach with recent state-of-the-art methods. Our approach achieves competitive performance, exhibiting the best ACER under three out of the four evaluation protocols. Notably, We compare our results with the noise-based methods to verify the effectiveness of our de-spoofing method. Specifically, we outperform the de-spoofing methods of FacesDs~\cite{jourabloo2018face} and Identity-DS~\cite{xu2021identity} in all protocols, while outperforming the SpoofTrace~\cite{liu2020disentangling} in three protocols.

\emph{3)Result on SiW:}
Given the attainment of 0\% ACER on Protocols 1 and 2 of the SiW dataset by recent methods, our comparison concentrates on evaluating the performance on Protocol 3. \tabref{tab:OULU-NPU} compares the performance of our method with four state-of-the-art methods, and we get comparable results on this protocol.

\begin{figure*}[t]
\begin{center}
   \includegraphics[width=0.9\linewidth]{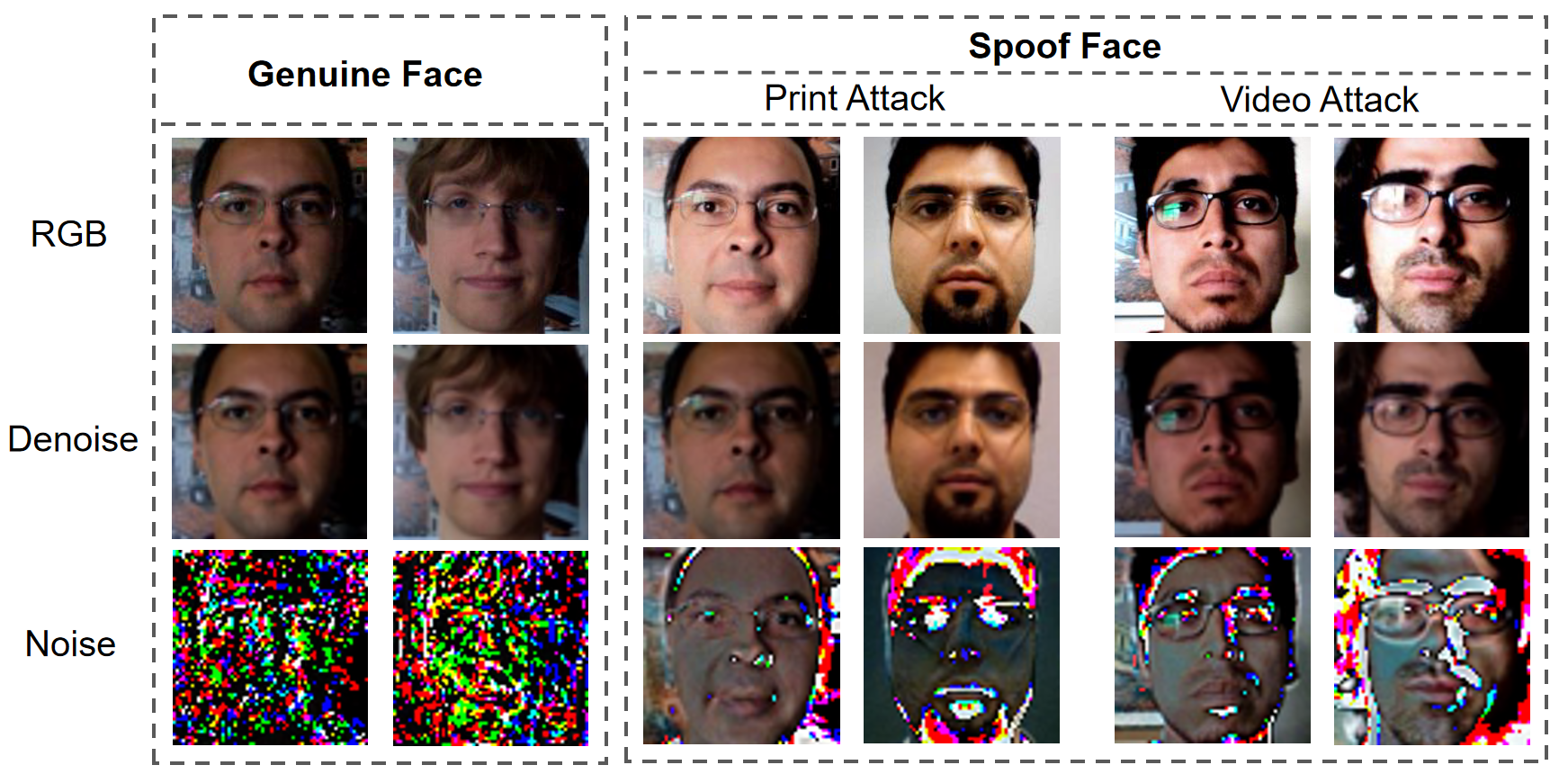}
\end{center}
   \caption{The de-spoofing results of genuine and spoofing faces are shown. The noise patterns extracted from the spoof faces are noticeably more prominent than those extracted from genuine faces. In contrast, the noise patterns extracted from genuine faces are more similar to the random noise error that occurs during the inverse process.}
\label{fig:Diff}
\end{figure*}

\subsection{Inter-Dataset Testing}\label{Inter}
Cross-testing aims to justify the generalization ability of the concerned model. We conduct the C2R and R2C benchmarks to manifest the effectiveness of our de-spoofing method. The first C2R protocol is trained on CASIA-MFSD and tested on Replay-Attack, while the R2C protocol is trained on Replay-Attack and tested on CASIA-MFSD. To ensure a fair comparison, we only compare with the methods that use a single frame for testing. It can be seen from~\tabref{tab:RC} that our proposed method achieves an HTER of 9.8$\%$ on C2R and 34.4$\%$ on R2C and outperforms the prior state-of-the art over 35.95$\%$ on the C2R protocol. In addition, our method shows superior average performance compared to prior approaches.

\begin{table}
    \begin{center}
        \caption{The results of inter-testing on C2R and R2C protocols. The evaluation metric is HTER($\%$). Lower HTER and Average are better.}
        \renewcommand\arraystretch{1.2}
        \label{tab:RC}
        \resizebox{\linewidth}{!}{
        \begin{tabular}{|c|c|c|c|c|c|} 
            \hline 
            \multicolumn{1}{|c|}{\multirow{2}{*}{Method}} & Train & Test & Train & Test &  \multicolumn{1}{c|}{\multirow{2}{*}{Average$\downarrow$}}  \\ \cline{2-5}
            & \tabincell{l}{CASIA- \\ MFSD} & \tabincell{l}{Replay \\ Attack} & \tabincell{l}{Replay \\ Attack} &  \tabincell{l}{CASIA- \\ MFSD}  &\\
            \hline
            CNN~\cite{yang2014learn} & \multicolumn{2}{c|}{48.5} & \multicolumn{2}{c|}{45.5} & 47.0\\
            LBP~\cite{boulkenafet2015face} & \multicolumn{2}{c|}{47} & \multicolumn{2}{c|}{39.6} & 43.3\\
            Identity-DS~\cite{xu2021identity} & \multicolumn{2}{c|}{27.1} & \multicolumn{2}{c|}{31.4} & 29.3\\
            FaceDs~\cite{jourabloo2018face} & \multicolumn{2}{c|}{28.5} & \multicolumn{2}{c|}{41.1} & 34.8\\
            BCN~\cite{yu2020face} & \multicolumn{2}{c|}{16.6} & \multicolumn{2}{c|}{36.6}  & 26.6\\
            Attention~\cite{chen2019attention} & \multicolumn{2}{c|}{30.0} & \multicolumn{2}{c|}{33.4} & 31.7\\
            CDCN~\cite{yu2020searching} & \multicolumn{2}{c|}{15.5} & \multicolumn{2}{c|}{32.6} & 24.1\\
            MEGC~\cite{bian2022learning} & \multicolumn{2}{c|}{20.2} & \multicolumn{2}{c|}{27.9} & 24.1\\
            DANet~\cite{sun2022danet} & \multicolumn{2}{c|}{18.0} & \multicolumn{2}{c|}{\textbf{27.6}} & 22.8\\
            SDCC~\cite{zhang2021structure} & \multicolumn{2}{c|}{15.3} & \multicolumn{2}{c|}{29.4} & 22.4\\
            \textbf{Ours} & \multicolumn{2}{c|}{\textbf{9.8}} & \multicolumn{2}{c|}{34.4} & \textbf{22.1} \\      
            \hline
        \end{tabular}
    }
    \end{center}
\end{table}

\subsection{Performance Analysis}\label{Analysis}
In order to provide additional evidence supporting the effectiveness of our De-spoofing Diffusion Model, we present the de-spoofing results of genuine and spoof samples in ~\figref{fig:Diff}. Through the transfer effect of the diffusion model, the noise patterns are separated effectively which enhances the performance of face anti-spoofing. The results show that the noise patterns extracted from the spoof samples are more prominent than those from the genuine samples. On the other hand, the noise patterns of genuine faces are similar to the random noise error in the inverse
process of DDIMs. This observation confirms the effectiveness of our model in identifying the presence of spoof attacks by isolating the noise patterns from the input images.

\section{Conclusion}\label{Conclusion}
This work proposes a novel approach to face anti-spoofing by designing a de-spoofing diffusion model. The diffusion model is leveraged to separate the noise patterns present in spoof images and recover the corresponding genuine images. To further exploit the detection cues in the spoof noise, a two-stream network is designed to fuse the RGB input and noise pattern, providing a more reliable and accurate method of face anti-spoofing. Extensive experiments are conducted to evaluate the proposed method, demonstrating its superiority over state-of-the-art methods across multiple evaluation protocols, including intra-testing protocols and inter-testing benchmarks. These results highlight the potential of diffusion models in the field of face anti-spoofing and emphasize the importance of enhancing the security and reliability of face recognition systems.

\section*{Acknowledgement}
This work was supported in part by Chinese National Natural Science Foundation Projects (\#62276254, \#62176256, \#61976229, \#62106264, \#62206280, and \#U2003111), the Youth Innovation Promotion Association CAS (\#Y2021131), the Defense Industrial Technology Development Program (\#JCKY2021906A001), and the InnoHK program.

{\small
\bibliographystyle{ieee}
\bibliography{egbib}
}

\end{document}